\documentclass{article}

\usepackage{PRIMEarxiv}

\usepackage[utf8]{inputenc} 
\usepackage[T1]{fontenc}    
\usepackage{hyperref}       
\usepackage{url}            
\usepackage{booktabs}       
\usepackage{amsfonts}       
\usepackage{nicefrac}       
\usepackage{microtype}      
\usepackage{lipsum}
\usepackage{fancyhdr}       
\usepackage{graphicx}       
\graphicspath{{media/}}     
\usepackage{multirow}
\usepackage{biblatex}
\usepackage{graphicx}
\usepackage{float}
\usepackage{amsmath}

\bibliography{references}  
\title{Semantic Information for Object Detection}

\author{
  Jean-Francois Nies \\
  RPTU Kaiserslautern-Landau\\
  Gottlieb-Daimler str. 47, 67663\\
  Kaiserslautern, Germany \\
  \texttt{j\_nies15@cs.uni-kl.de} }

\begin{document}
\maketitle

\begin{abstract}
  In this paper, we demonstrate that the concept of \textit{Semantic Consistency} and the ensuing method of \textit{Knowledge-Aware Re-Optimization} can be adapted for the problem of object detection in intricate traffic scenes. Furthermore, we introduce a novel method for extracting a knowledge graph from a dataset of images provided with instance-level annotations, and integrate this new knowledge graph with the existing semantic consistency model.

  Combining both this novel 'hybrid' knowledge graph and the pre-existing methods of frequency analysis and external knowledge graph as sources for semantic information, we investigate the effectiveness of knowledge-aware re-optimization on the Faster-RCNN and DETR object detection models. We find that limited but consistent improvements in precision and/or recall can be achieved using this method for all combinations of model and method studied.
\end{abstract}

\section{Introduction}

The problem of \textit{Object Detection}~\cite{Zou2019} may be expressed informally as the question "\textit{What objects are where?"}. More formally, it incorporates both the definition of bounding boxes encompassing objects of interest as closely as possible, and the ability to assign correct labels to each such box.  In the last years, research in this domain, chiefly using convolutional neural networks (CNN), has already led to object recognition systems capable of surpassing human performance for selected tasks~\cite{Altenberger2018}. Such systems typically employ deep CNNs and are widely deployed in a range of practical applications in areas such as autonomous vehicles and facial recognition systems.

However, unlike a human observer, a model such as a convolutional neural network has no ability to reason about the scene it is observing, and dismiss or correct erroneus identifications based on context \cite{lecun2015deep}. Other objects present, the time and location, and the position of objects in relation to each other may all provide vital cues to an onlooker, but the integration of this context into a machine learning model is a complex problem: Contextual information must be made available in a machine-readable form, frequently resulting in large amounts of additional data, and processed alongside conventional inputs. Nonetheless, this approach has received a great deal of attention. Aside from immediate gains in performance, such \textit{Informed Machine Learning} may also provide 'soft' benefits in terms of explainability and accountability.~\cite{Rueden2019}

While it is possible to design models which integrate external knowledge from the ground up, there may also be situations in which the necessary training data is not available, or where a proprietary model must be treated as a 'black box'. In either case, the ability to perform knowledge-based re-optimization of the outputs of arbitrary models may be desirable.

In this report, we investigate one variant of this approach, using Semantic Consistency Matrices. Basing our work on an implementation by Lemens et al. \cite{Lemmens2021} of the algorithm described by Fang et al. \cite{Fang2017}, we evaluate the effectiveness of this method on the Cityscapes dataset, and introduce an additional method for generating semantic consistency matrices which acts as a compromise between the already existing methods. Furthermore, we study the effectiveness of the different knowledge-aware re-optimization methods on the performance of the more recent Deformable Transformer (DETR) architecture.

\section{Related Work}
Object detection is an active field of research with numerous immediate applications. It should therefore come as no surprise that several different approaches are currently under active investigation. 

Convolutional neural networks remain a fundamental part of object detection models, but many recent models augment their capabilities by introducing different additional mechanisms. For example, the Faster R-CNN model \cite{Ren2016} is widely used as a backbone for other object detection models. It makes use of an RPN or region proposal network to identify regions of interest before applying a convolutional neural network to the actual task of recognition on these regions. By contrast, the DETR architecture \cite{Carion2020} prepends its own CNN backbone to another model, composed of encoding and decoding transformers, such that the CNN acts as a dimensionality reduction mechanism.

Finally, the YOLO family of models \cite{redmon2018yolov3} typifies the one-shot object detection approach, named in contrast to the two stages of an FRCNN-like detector with separate region proposal and classification stages. A YOLO-style model, as indicated by expanding the acronym to "You Only Look Once", instead uses a single stage.

All these models, however, suffer from a limitation when compared to human perception: While they are able to extract large amounts of relevant information from input images, they do not generally leverage external, contextual knowledge such as causal or semantic relations between different depicted objects (whereby, e.g., an image containing cars is more likely to also contain an omnibus than a lobster) and image metadata. Given the intuitive nature of this approach, it should come as no surprise that the question of using additional knowledge in object detection has been considered from many different angles beyond the one at the core of this paper.

Liu et al.(2021)~\cite{Liu2021} use a \textit{similarity network} to measure the pairwise semantic similarity between objects, in a method somewhat similar to the one discussed in this report. However, unlike semantic consistency, this is not related to the likelihood of co-occurence of concepts. Instead, semantic similarity is a measure of how likely any two objects are to belong to the same class, no matter which class this may actually be.  Others, such as Zhu et al.~\cite{Zhu2021SemanticRR} have put forward architectures capable of more sophisticated integration not only of semantic, but also spatial information to improve the performance of object detection models. Chen et al.~\cite{Chen2020} demonstrate that such an approach can yield improved performance especially on small objects.

Menglong et al. (2019)~\cite{Menglong2019} applied a related approach to the task of image classification or object recognition. For this, they described a method in which labels assigned by one or several existing classifiers are used to construct a knowledge graph, in which semantic similarity between any two categories is assigned on the basis of the frequency at which these classifiers confuse them. This information is then used to refine image classification by adjusting the classifier's output to better reflect the domain knowledge gathered, assigning greater or lower confidence to outputs according to their plausibility.

The construction of knowledge graphs is moving beyond simple text databases, and into the field of Multi-Model Knowledge Graphs, organically integrating image and text data into a single structure and explicitly connecting images to explicit visual properties, as in Zhu et al.~\cite{zhu2022multi}.

Making use of the fact that many images available today are tagged with temporal and geographic metadata, MacAodha et al.~\cite{MacAodha2019PresenceOnlyGP} take a different direction and extend conventional object detection models to consider this metadata as additional context. This is accomplished through the use of an assessment of co-occurrence of objects in given times and places.

Von Rueden et al.~\cite{Rueden2019} introduce the notion of 'informed machine learning' as an umbrella term for all types of machine learning making use of external information sources, and present a useful general taxonomy.

Finally, Castellano et al.~\cite{castellano2022leveraging} demonstrate a novel area of application of these techniques, by applying a combination of knowledge graphs and deep learning techniques to the analysis of artworks, and provide a related knowledge graph.

\section{Dataset}
We evaluated the Knowledge-Aware Re-Optimization method on the Cityscapes Dataset~\cite{Cordts2016}, which was originally designed for scene understanding as applied to urban scenes. It contains 5000 stereoscopic images with detailed annotations at the instance and pixel levels, as well as additional annotations with coarser annotations. In our case, we consider the instance-level annotations. 

In order to include depth information necessary for some applications, the dataset contains pairs of stereoscopic images, i.e. pairs of images recorded simultaneously by two cameras placed at a slight distance from each other but facing the same direction. The left- and right-hand image thus differ slightly, as the perspective of the left and right eye would differ in human vision, enabling depth perception. As depth information was outside the scope of our experiments, we followed the convention adopted by the implementation by T. Beemelmanns~\cite{Beemelamanns2022}for dealing with cityscapes dataset formatting: Of each pair of stereoscopic images, only the left-hand one was retained, thus providing us with a dataset of single images.

\begin{center}
    \begin{table}[]
        \centering
        \begin{tabular}{|c| c c|}
        \hline
        Class & Training & Validation\\
        \hline
        Person & 12044 (33.1\%)& 3450 (34.4\%)\\
        Rider & 1080 (2.9\%)& 542 (5.4\%)\\
        Car & 19113 (52.6\%)& 4378 (43.7\%)\\
        Truck & 312 (0.8\%)& 93 (0.9\%) \\
        Bus & 245 (0.6\%)& 98 (0.9\%) \\
        Train & 118 (0.3\%)& 23 (0.2\%) \\
        Motorcycle & 492 (1.3\%)& 148 (1.4\%)\\
        Bicycle & 2903 (7.9\%)& 1281 (12.7\%) \\
        Total & 36307 & 10013 \\

        \hline
        
        \end{tabular} 
        \bigskip
        \caption{ Distribution of classes across the partial Cityscapes dataset}
        \label{dataset}
    \end{table}
\end{center}

The number of instances in the training and validation components of the dataset is given in table \ref{dataset}. This table also illustrates an issue of the dataset, namely the large class imbalance present across both the training and validation sets. Thus, the number of e.g. pedestrians (designated as 'Person' in the dataset label but distinct from a 'Rider' on a bicycle or motorcycle) in the validation set exceeds the number of trains or busses by two orders of magnitude. The classes corresponding to Trucks, Buses and Trains are especially underrepresented. However, the proportion of each class relative to the total number of instances in both sets is consistent except for the classes 'Car' and 'Bicycle', where the training and validation sets differ from each other by 5 percentage points.

\section{Methodology}
In this paper, we adapt the concept of \textit{semantic consistency}~\cite{Fang2017} for the domain of autonomous driving to improve the performance of object detection models in traffic scenes by using environmental knowledge. This knowledge-aware re-optimization framework relies on a semantic consistency matrix. For any two concepts, this matrix gives a semantic consistency value, which is an indication of how likely instances of these two concepts are to occur simultaneously in an image. This includes the semantic consistency of a concept with itself, as multiple instances of the same concept in a single image are more likely for some concepts than others.

The framework as such is independent of any particular object detection method or implementation, instead treating the object detection model as a black box. The model's output is adjusted to better fit the semantic consistency matrix, with the scores for each class being raised or lowered to increase the overall consistency of the detection. To regularize the model and avoid being overly restricted in cases which do not conform to the expected distribution, very large changes in class scores are also penalized. For our experiments, we employed {FRCNN}~\cite{Ren2016} and DETR~\cite{Carion2020} as our object detection model.

The main challenge for this approach lies in obtaining the semantic consistency matrix, $S$. We investigated three different methods, two from the original paper of Fang et al.\cite{Fang2017} on re-optimization and one novel hybrid method.

\subsection{External Knowledge Sources}
\subsubsection{Frequency-Based Semantic Consistency}
The frequency-based method creates a semantic consistency matrix directly from the annotated training data also used for the backbone model, and therefore requires this training set to be available. Fang et al.\cite{Fang2017} note that this may be a drawback in practical applications where this training data may be proprietary or otherwise unavailable, and also report a lower performance for this method. Conversely, however, it may be useful for situations where no suitable knowledge graph is available.
The frequency-based method for determining semantic consistency is based on co-occurrences of concepts. Thus, it is assumed that objects which frequently occur together in the training set have a higher degree of semantic consistency, following the equation

\[S_{l,l'}=max(log \frac{n(l,l')N}{(n(l)n(l'))},0)\]

where $n(l)$ and $n(l')$ are the individual frequencies of the respective concepts $l,l'$, $n(l,l')$ denotes the number of co-occurrences of the two concepts and $N$ is the total number of instances. As the number of co-occurrences of multiple instances of one and the same class may also be relevant, a special 'handshake' equation is used in this case, following the same conventions:

\[n(l,l)=n(l)\frac{n(l)-1}{2}\]

\subsubsection{Knowledge Graph-Based Semantic Consistency}
Knowledge Graph-Based Semantic Consistency is the second method described by Fang et al.\cite{Fang2017} It is capable of being used even on pre-trained models where no training data is available, and can give better performance than the frequency-based method. 

To this end, an external knowledge graph whose vertices represent different concepts, and whose edges connect the semantically related concepts is required. For processing the knowledge graph, we followed the implementation of Lemmens et al.~\cite{Lemmens2021} , who provided some key clarifications to the original paper. Their method involves two stages.

First, the pre-existing graph, in our case ConceptNet5~\cite{Speer2012}, is cropped to include only positive relations (i.e., remove relations such as antonyms and contradictions) and is limited, without loss of generality, to the English-language versions of concepts.

Then, the semantic consistency matrix is derived from this knowledge graph using a series of random walks starting from each concept of interest, i.e. corresponding to a desired label of the object detection model. The random walk then traverses the graph, but also has a low probability of resetting to the original node with every step to avoid remaining stuck in local groups (Random Walk with Restart, RWR)~\cite{Tong2006} The probability of reaching any given other node from the starting node eventually converges towards a steady-state after a sufficient number of iterations.

\begin{figure}
    \centering
    \includegraphics[width=.9\textwidth]{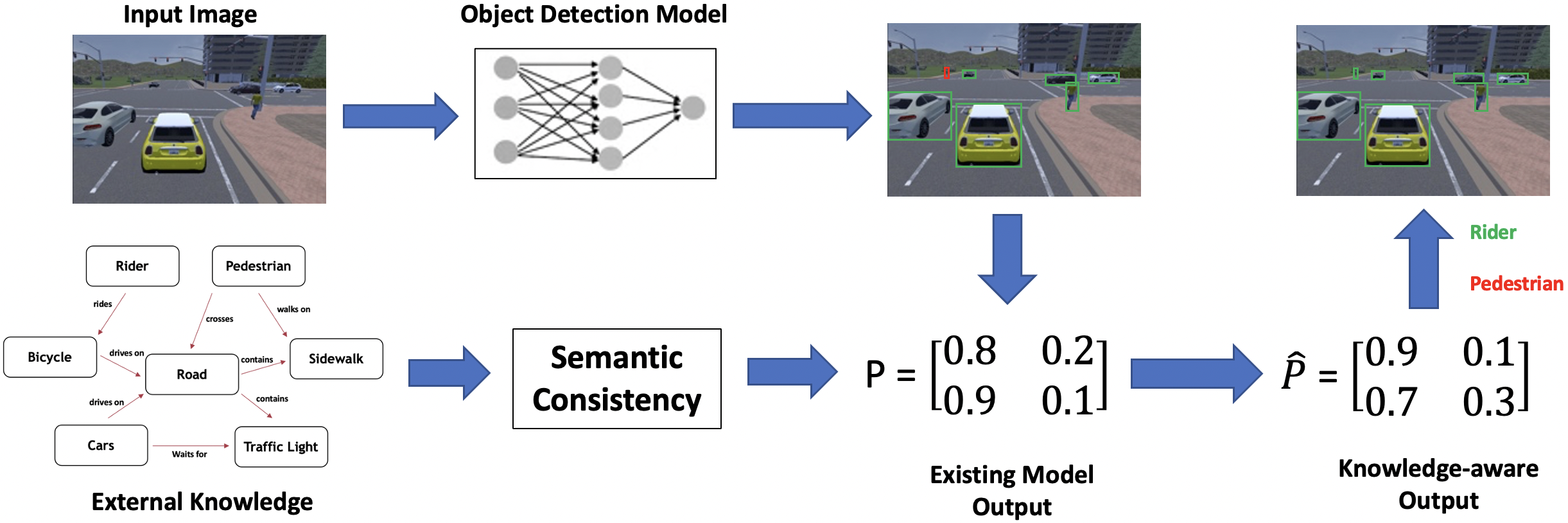}
    \caption{The basic principle of knowledge-based re-optimization. Using a semantic consistency matrix, the model output $p$ is changed to a re-optimized output $\hat{p}$}
    \label{fig:pipeline_overview}
\end{figure}

\subsubsection{Hybrid Semantic Consistency}
In this paper, we introduce another method for knowledge-based re-optimization, derived from a combination of the two approaches described above. Intuitively, the semantic consistency as derived from an external knowledge graph will, for a well-designed graph, correspond to the degree to which these concepts are related in the corpus the graph is based on. However, this may not be ideal in every situation. If a model is to be applied in a specific use-case instead of general-purpose object detection, the class distribution it encounters may be different from that suggested by a consistency matrix derived from a generic knowledge-graph. As a simple example, a knowledge graph derived from an encyclopedia might contain relatively few mentions of pedestrians, but greater emphasis on trains, whereas an object detection model intended for self-driving vehicles is likely to encounter the opposite situation.

However, as the knowledge graph-based method produced performance superior to the frequency-based method overall, we investigated the possibility of creating a knowledge graph tailored to our dataset based on frequency data. Our assumption is that a well-constructed dataset can be more relevant to the area of application of the model than a general knowledge model. Our approach uses the following steps:
\begin{itemize}
    \item We generate a matrix $M_1$ of co-occurrences for each class across our dataset.
    \item We generate a knowledge graph $G$ based on $M_1$ by creating a vertex and concept for each label present in the co-occurrence matrix, and connect them when the corresponding number of co-occurrences exceeds a threshold $\gamma$.
    \item A random walk is performed on this new graph $G$, and the semantic consistency matrix is generated as in the case of ordinary graph-based semantic consistency.
\end{itemize}

We thus arrive at a semantic consistency matrix with the same structure to that produced by other methods.

\subsection{Knowledge-Based Re-optimization}

Having generated our semantic consistency matrix $S$, we now perform the actual re-optimization. This is done by finding the minimum of a loss function accounting for both semantic consistency and the original input.

Formally, given two bounding boxes $b,b'$, we call $P_{b,l},P_{b',l'}$ the probability returned by the model for a label $l$ resp. $l'$ for either box. The loss function to be minimized is given by the equation below, where $L$ is the number of concepts and $B$ number of bounding boxes.

\begin{align}
    \label{loss1}
    E(\hat{P}) = (1-\epsilon)\sum^{B}_{b=1}\sum^{B}_{b'=1,b'\neq b}
    \sum^{L}_{l=1}\sum^{L}_{l'=1} S_{l,l'} (\hat{P}_{b,l}-\hat{P}_{b',l'})^2 \\+
    \label{loss1a}
    \epsilon \sum^{B}_{b=1}\sum^{L}_{l=1} B S_{l,*} (\hat{P}_{b,l}-P_{b,l})^2
\end{align}

This function can be decomposed into a sum of two terms, weighted by the parameter $\epsilon \in (0,1)$, which must be determined in practice for each dataset. The first term in equation \ref{loss1} effectively demands that the total square error between the final output for semantically similar concepts be minimized. The semantic consistency acts like a weight for the nested sum operations, meaning that differences between wholly orthogonal concepts ($S_{l,l'}=0$) will have no effect at all on the final value.

The second term, in equation \ref{loss1a}, imposes a cost for overly large deviations from the backbone model's output. The square error between the final output $\hat{P}$ and the model's output $P$ is modified by a coefficient and added for all classes and bounding boxes, effectively acting as a check on the re-optimization. The cost function given in \ref{loss1} is then minimized. This is accomplished by setting the gradient of $E(\hat{P})$ w.r.t. $\hat{P}_{b,l}$ to zero.

\subsection{Metrics}
The implementation of Lemmens et al.~\cite{Lemmens2021} made use of an Area-Under-Curve (AUC) style of average precision calculation, computing class-wise average precision for a range of recall thresholds running from $0$ to $1.0$. We retained this system, as it is especially useful for automatic parameter searches, where it implicitly balances the priority of precision and recall even if only the former is explicitly targeted.

\subsection{Experiment Structure}

As a first step, we established a baseline of performance metrics by measuring the performance of both architectures on the validation set of the cityscapes dataset. We then attempted to find an optimal combination of re-otimization hyperparameters. As there are several such parameters, we resorted to an automatic parameter search using Optuna \cite{optuna_2019}, covering the weight parametre $\epsilon$ discussed above, but also the number of adjacent boxes and classes considered for re-optimization, and an internal score threshold for re-optimized outputs. As the precision-recall curve metrics used already encompassed recall in computing precision, we verified that choosing accuracy as the sole optimization target had no negative effect and carried out our experiments on this base.

Tests showed that the best results on the unoptimized baseline were obtained when considering the 100 highest-scoring detections. As this criterion was also used in the original paper \cite{Fang2017}, we opted to follow this convention. However, we also considered an alternative method, using a score threshold rather than a static number of detections, and the use of different values for both number of detections and threshold. The results are detailed in section \ref{results}.

\subsection{Challenges}

 Lemmens et al. \cite{Lemmens2021} have already pointed out some technical difficulties which hindered their re-implementation based on the original paper. These difficulties arose primarily from missing details, such as the exact stage of the pipeline at which the top-scoring detections are selected. During our investigation, we also observed that, while the re-optimization model is mathematically capable of treating the backbone model as a black box, practical implementations required additional care.

While investigating different state-of-the-art, or recent implementations, we found that many object detection frameworks differ in terms of input formats, output formats, or both, in a way which complicated integration with knowledge-based re-optimization even where an explicit API is provided. As an example, yolov3 \cite{redmon2018yolov3}, as implemented in the mmdetection framework, requires image input to be given as a file path, and returns a list of bounding boxes and confidence scores per class, while the implementation of Faster R-CNN built into pytorch requires image inputs in the form of pytorch tensors, and returns a dictionary holding labels, scores and bounding boxes in continuous tensors. It is therefore necessary to implement explicit format conversions for each architecture.

An unrelated but significant difficulty stemmed from the large imbalance of classes within the cityscapes dataset. While FRCNN was not visibly affected by this, we found that DETR suffered from degraded performance on the underrepresented classes (see section \ref{sec_detr}). Based on the findings discussed below, it seems probable that a larger, more balanced training set or data augmentation focused on the least-represented classes would have greatly strengthened the results achieved with DETR.

\section{Results}
\label{results}

\subsection{Faster-RCNN}

In this section, we present and analyze the key results obtained with the Faster R-CNN architecture. These results were computed on the 100 top-scoring detections, and are given for the best-performing configuration of hyperparameters for the given experiment.

\begin{figure}
    \centering
    \includegraphics[width=.55\textwidth]{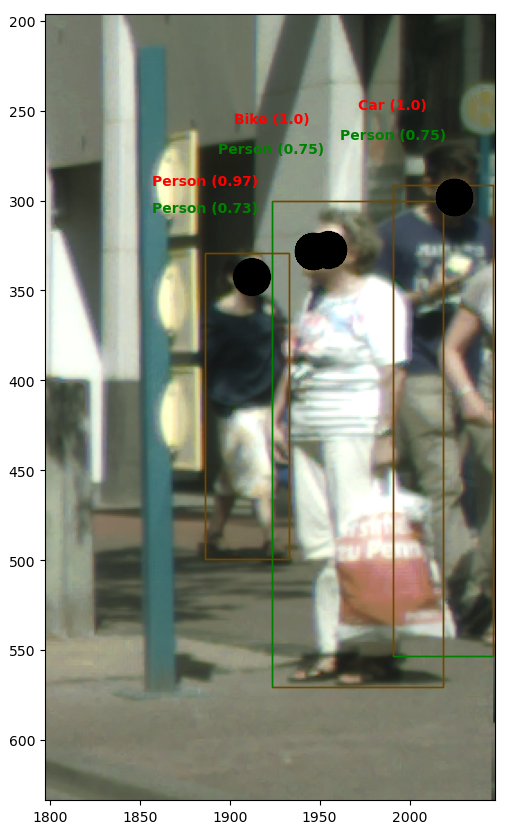}
    \includegraphics[width=.90\textwidth]{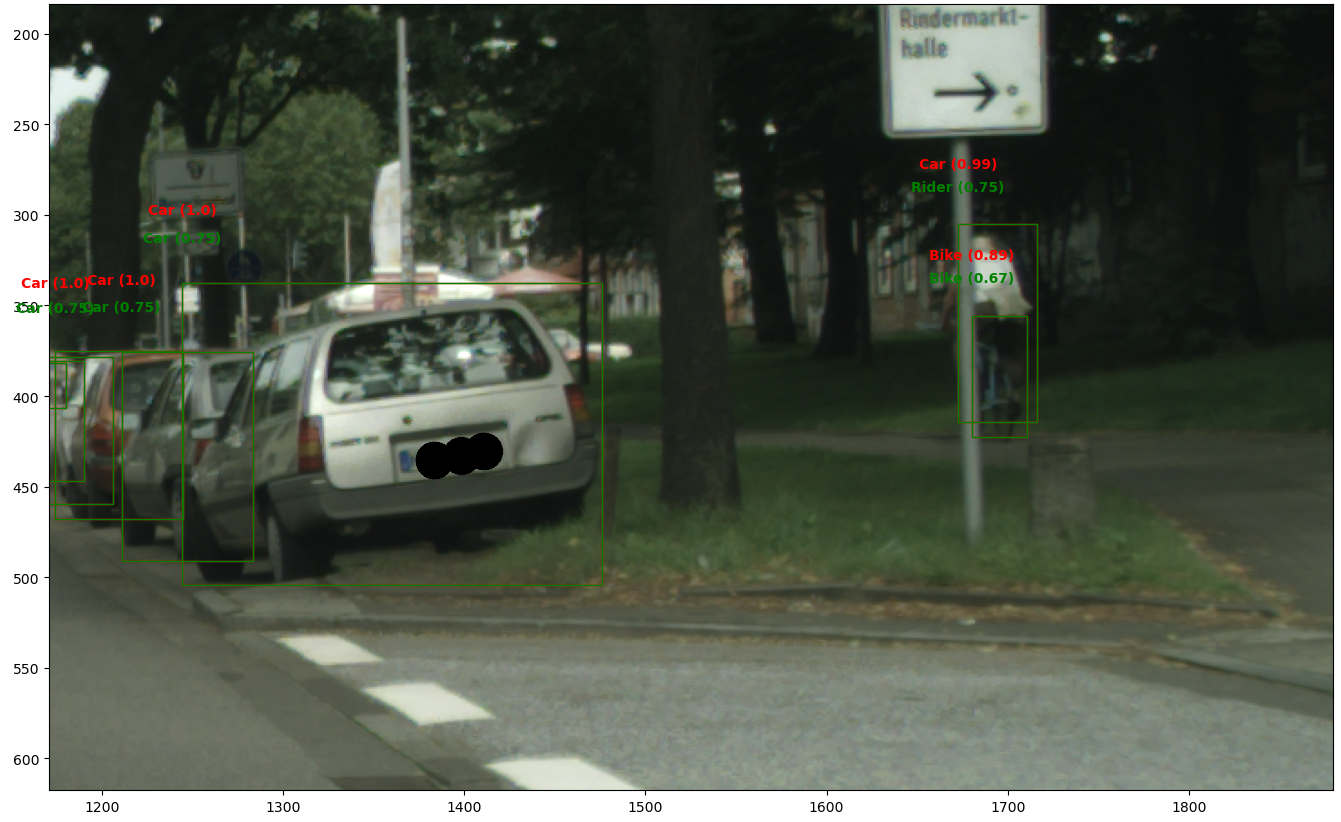}
    \caption{Examples of un-optimized output (red) being adjusted to a re-optimized label (green), with the corresponding confidence score. (Faces and license plates redacted manually)}
    \label{fig:my_label}
\end{figure}

\begin{figure}
    \centering
    \includegraphics[width=0.90\textwidth]{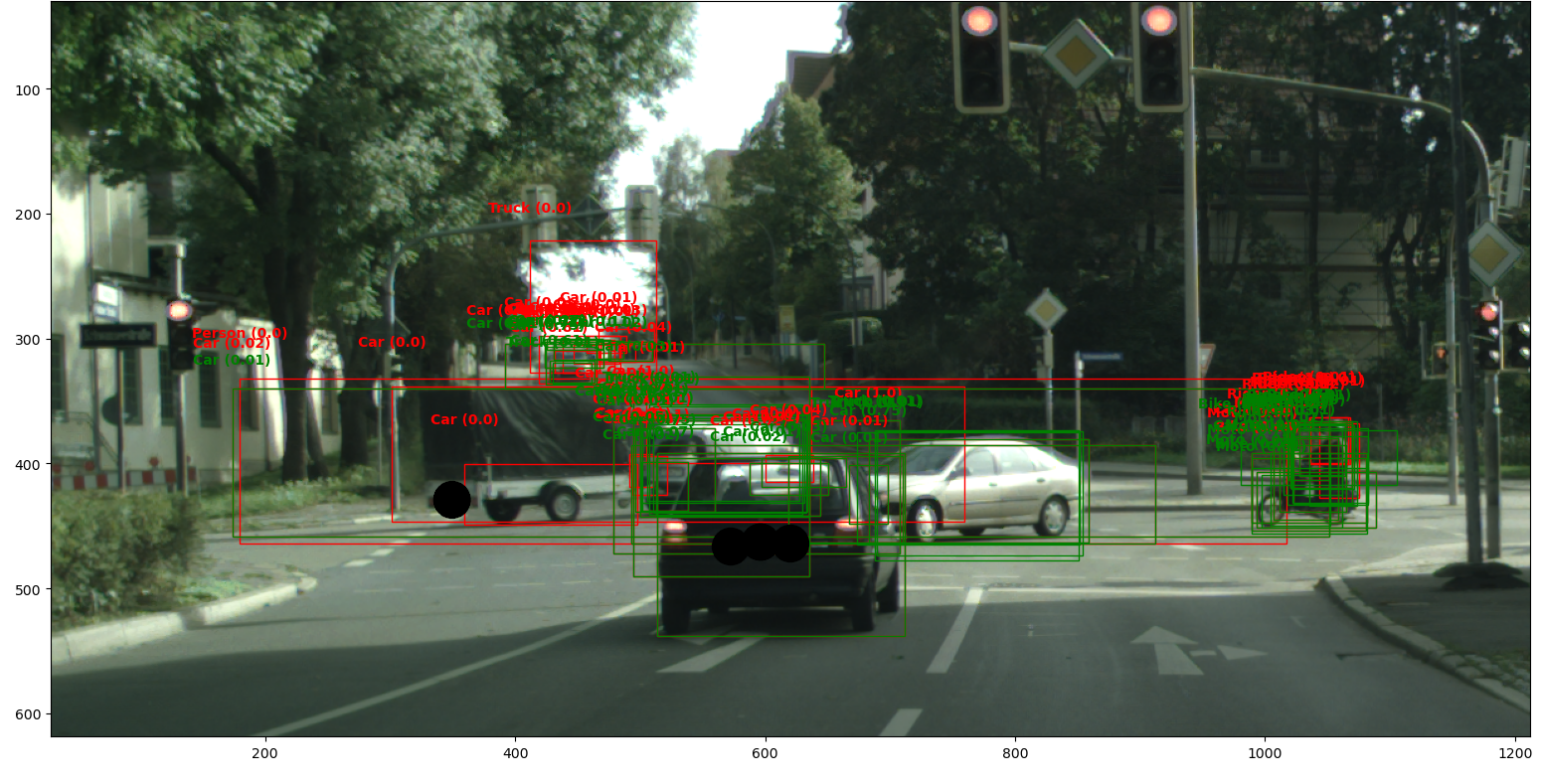}
    \caption{Example of spurious detections in the un-optimized output (red) being removed in the re-optimized output (green), with the corresponding confidence score. }
    \label{fig:my_label2}
\end{figure}

\begin{table}[]
    \centering
    \begin{tabular}{|c||c|c|c|c|c|}
        \hline
        Re-Op     &  mAP@100   & Recall@100 & Small & Medium & Large \\
        \hline
        Baseline  &  25.35         & 35.60         & 12.26 & 34.77 & 57.43 \\
        Hybrid    &  25.01 (-0.34) & 35.87 (+0.27) & 11.94 & 35.00 & 58.40 \\
        KG        &  25.22 (-0.13) & 35.88 (+0.28) & 11.93 & 35.40 & 57.93 \\
        Freq.     &  25.40 (+0.04) & 35.84 (+0.24) & 12.21 & 35.11 & 57.80\\
        \hline
        
    \end{tabular}
    \bigskip
    \caption{Effects of different re-optimization methods on the FRCNN model's performance. Numbers in bracket give the difference from the baseline.}
    \label{tab:frcnn}
\end{table}

All methods, as shown in table \ref{tab:frcnn}, provided some gain in recall, with the largest gains being made using the knowledge graph-based semantic consistency matrix. However, the knowledge-graph (KG) method also resulted in some decrease in mAP. The hybrid method fared worse in both respects, with smaller recall gains and greater mAP losses. Finally, the frequency-based method resulted in an increase in recall slightly smaller than the previous methods, but also no loss (and, in fact, a marginal gain) of mAP. This is broadly consistent with the results of Lemmens et al.\cite{Lemmens2021}: Although reported an decrease in mAP and an increase in recall with both knowledge- and frequency-based methods on MSCOCO, the pattern is similar to the extent that the knowledge-based method resulted in greater reported gains and losses respectively. The effects of the re-optimization, both positive and negative, reported by Lemmens et al. are on the same order of magnitude as ours, with the overall effect not exceeding 1\%. This is substantially less than what was originally reported in the paper of Fang et al. \cite{Fang2017}, where recall gains reported could reach 3\%.

\begin{table}[]
    \label{classwise1}
    \centering
    \begin{tabular}{|c| c c c c |c c c c|}
         \hline
         \multirow{2}{*}{Method} & \multicolumn{4}{c|}{AP} & \multicolumn{4}{c|}{Recall} \\
         \cline{2-4} \cline{5-9}
                        & Baseline & KG & Hybrid & Freq & Baseline & KG & Hybrid  & Freq \\
         \hline
         Person         & 25.29 & 24.92 & 24.42 & 25.32 & 33.51 & 33.78 & 33.94 & 33.48 \\
         Rider          & 29.76 & 29.40 & 29.73 & 29.75 & 41.31 & 40.26 & 40.18 & 41.25\\
         Car            & 44.28 & 43.74 & 42.73 & 44.26 & 49.87 & 50.67 & 50.76 & 49.82\\
         Truck          & 19.36 & 19.45 & 19.46 & 19.49 & 31.51 & 32.58 & 32.37 & 32.69\\
         Bus            & 36.83 & 36.86 & 36.90 & 36.86 & 46.37 & 46.84 & 46.63 & 46.73\\
         Train          & 12.14 & 12.15 & 11.82 & 12.15 & 23.48 & 23.91 & 23.91 & 23.91\\
         Motorcycle     & 14.87 & 15.00 & 15.11 & 15.04 & 27.18 & 27.65 & 27.99 & 27.72\\
         Bicycle        & 20.28 & 20.26 & 19.95 & 20.33 & 31.21 & 31.35 & 31.19 & 31.14\\
         \hline
         
    \end{tabular}
    \bigskip
    \caption{Class-wise breakdown of Precision and Recall using Hybrid Re-Optimization with an FRCNN backbone}
    \label{frcnn}
\end{table}

Examining the results in greater detail, we find that the average values discussed in the earlier do not arise from a homogeneous change to all classes in the same direction. The knowledge graph-based and hybrid methods do give poorer mAP values on some classes, but actually yield increased precision on others. Specifically, it appears that the classes making up a smaller percentage of the dataset, such as buses and motorcycles, do benefit from this method, whereas precision scores for the classes with the greatest support in the dataset, Person and Car, also show the greatest decrease in precision. The frequency-based method was more consistent, with increases in precision across the board, with the exception of the Car class. The latter effect may be tentatively explained the fact that this class makes up a substantially larger portion of the training set (53.6 \%), from which the frequency-based consistency matrix is derived, than of the validation set (43.7\%), on which it is applied.

\subsection{Deformable Transformers}
\label{sec_detr}

In this section, we consider the results achieved with the different re-optimization methods applied to an implementation of the DETR architecture \cite{Carion2020} on the Cityscapes dataset. As before, we will focus primarily on metrics computed for the 100 highest-scoring detections.

Performance for DETR was typically lower than for FRCNN, primarily due to slightly lower recall. However, the effects of the knowledge-aware re-optimization largely followed the same pattern.

\begin{table}[]
    \centering
    \begin{tabular}{|c||c|c|c|c|c|}
        \hline
        Method     &  mAP@100   & Recall & Small & Medium & Large \\
        \hline
        Baseline  & 21.84 & 35.48 & 17.46 & 44.63 & 62.71 \\ 
        Hybrid    & 21.81 (-0.033) & 35.61 (+0.13) & 17.46 & 44.71 & 63.03 \\
        KG        & 21.81 (-0.031) & 35.63 (+0.15) & 17.47 & 44.71 & 63.18 \\ 
        Freq.     & 21.58 (-0.038) & 35.70 (+0.14) & 17.46 & 44.70 & 63.06 \\ 
        \hline
    \end{tabular}
    \bigskip
    \caption{Effects of different re-optimization methods on the DETR model's performance. Numbers in bracket give the difference from the baseline.}
    \label{tab:frcnn}
\end{table}

As with Faster R-CNN on Cityscapes and COCO\cite{Lemmens2021}, the same pattern as on FRCNN is reproduced, albeit weakly. One oddity of the results on DETR, which we discovered by accident and explored for the sake of completeness, lay in an unexpected behaviour of the $\epsilon$ hyperparameter. $\epsilon$  governs the relative weight of the semantic consistency values and the backbone's output. It is therefore normally constrained to a range of $0 < \epsilon < 1$. However, we found that that using a value of $\epsilon>1$ can in fact also lead to an increase in model performance for certain parameter combinations.  The anomalous results obtained with $\epsilon>1$, consistently display relatively large increases in precision at the expense of recall, which, combined with the weaker gains made by the ordinary ($\epsilon<1$) method, suggests that some backbone models may benefit more from treating the knowledge-aware component as a 'pessimizer' with an effectively negative weight. This effect did not occur with the Faster-RCNN architecture when we repeated the experiments using this architecture under the same conditions, with the overall change in performance being an order of magnitude smaller than the one achieved with the more orthodox approach. These results are included in Appendix 1.

\begin{table}[]
    \label{classwise1}
    \centering
    \begin{tabular}{|c| c c c c |c c c c|}
         \hline
         \multirow{2}{*}{Class/Method} & \multicolumn{4}{c||}{AP} & \multicolumn{4}{c|}{Recall} \\
         \cline{2-4} \cline{5-9}                   
                      & Baseline & KG & Hybrid & Freq & Baseline & KG & Hybrid  & Freq \\
         \hline
         Person       &  28.58  & 28.54 & 28.40 & 28.53 & 40.75 & 40.77 & 40.10 & 40.77\\
         Rider        &  27.53  & 27.59 & 27.74 & 27.59 & 37.22 & 37.24 & 36.94 & 37.23\\
         Car          &  49.03  & 48.95 & 49.01 & 48.95 & 57.66 & 57.71 & 57.23 & 57.71\\
         Truck        &  11.88  & 11.88 & 12.32 & 11.91 & 29.73 & 30.57 & 28.51 & 29.46\\
         Bus          &  17.13  & 17.04 & 18.33 & 17.05 & 29.25 & 29.42 & 29.40 & 29.39\\
         Train        &   2.52  &  2.55 &  3.94 &  2.53 & 23.77 & 23.84 & 23.19 & 23.84\\
         Motorcycle   &  14.70  & 14.62 & 15.44 & 14.61 & 26.76 & 26.76 & 26.74 & 26.76\\
         Bicycle      &  23.36  & 23.33 & 23.71 & 23.33 & 38.71 & 38.72 & 37.88 & 38.71\\
         \hline
         
    \end{tabular}
    \bigskip
    \caption{Class-wise breakdown of Precision and Recall using Hybrid Re-Optimization with a DETR backbone}
    \label{detr}
\end{table}

The apparent anomaly of the very low precision value for the 'Train' class may be explained by the fact that it is the least represented class in the training set with 118 instances, but especially in the validation set, where only 21 trains occur. Therefore, results for this class may not be significant. Conversely, when comparing the performance of Faster RCNNN in table \ref{frcnn} with that of DETR in table \ref{detr}, we find that DETR's precision and recall for the highest-scoring classes 'Person' and 'Car' match or surpass FRCNN's performance despite lower average values. As shown in table \ref{dataset}, these are also the most represented classes in the dataset. 

\subsection{Behaviour of Confidence Scores}

In view of the comparatively limited effect of re-optimization on model performance, we decided to investigate the way in which it modifies the confidence scores assigned by the backbone model in greater detail. The behaviour of the confidence scores, it was hoped, would help us to ensure that the model was functioning properly, but might also highlight differences in the behaviour of the FRCNN model as applied on the cityscapes dataset and the original MS-COCO dataset.

For each configuration in table \ref{scores}, we give descriptive statistics for the 100 results with the highest confidence scores (@100) for each image, consistent with the overall evaluation methodology. As may be seen in the table \ref{scores}, the results using the FRCNN architecture as a backbone are largely consistent between the two datasets, with some differences caused by different re-optimization parameters. The DETR backbone displays a tendency towards more extreme changes in score, but with an average change in confidence score remaining very close to zero overall. The fact that the standard deviation is larger than for FRCNN, but still far from the extreme values, suggests that the minima and maxima of confidence score changes for DETR are outliers rather than the rule.

\begin{table}[]
    \centering
    \begin{tabular}{|c c c|c c c c|}
        \hline
         Backbone & Dataset & Method & Mean Change & Maximum & Minimum & Std. Dev.\\
         \hline
         FRCNN & COCO & KG   & -0.14 & 0.04 & -0.25 & 0.08 \\
         FRCNN & COCO & Freq.& -0.14 & 0.03 & -0.25 & 0.08 \\
         FRCNN & Cityscape & KG     & -0.01 & 0.02 & -0.05 & 0.02 \\
         FRCNN & Cityscape & Freq.  & -0.02 & 0.06 & -0.1 & 0.03  \\
         FRCNN & Cityscape & Hybrid & -0.08 & 0.02 & -0.4 & 0.13  \\
         DETR & Cityscape  & KG     & 0.00  & 0.94 & -0.94 & 0.29  \\
         DETR & Cityscape  & Freq.  & -0.01 & 0.89 & -0.94 & 0.28 \\
         DETR & Cityscape &  Hybrid & -0.00 & 0.93 & -0.94 & 0.29 \\
         \hline
         
    \end{tabular}
    \bigskip
    \caption{Statistical analysis of changes in confidence scores for different datasets, backbone architectures and re-optimization methods.}
    \label{scores}
\end{table}

\section{Conclusion}
In this report, we investigated the possibility of using knowledge-aware re-optimization for object detection in on the Cityscapes dataset, applying the re-optimization model to a problem in the domain of autonomous driving. These experiments were carried out using three different sources of external knowledge, including a novel hybrid knowledge graph generated from frequency data obtained from the training set. We reported results for two different architectures.

Despite technical challenges in adapting the re-optimization model to different backbone architectures, we are able to apply the methods described in the original paper to the new configurations described above. We were able to achieve marginal but consistent increases in recall for both FRCNN and DETR on the cityscapes dataset with a favorable trade-off in precision.

\printbibliography

@Article{Chen2020,
  author    = {Chen, Shengjia and Li, Zhixin and Tang, Zhenjun},
  title     = {Relation R-CNN: A graph based relation-aware network for object detection},
  journal   = {IEEE Signal Processing Letters},
  year      = {2020},
  volume    = {27},
  pages     = {1680--1684},
  publisher = {IEEE},
}

@Article{Altenberger2018,
  author      = {Felix Altenberger and Claus Lenz},
  title       = {A Non-Technical Survey on Deep Convolutional Neural Network Architectures},
  date        = {2018-03-06},
  eprint      = {1803.02129v1},
  eprintclass = {cs.CV},
  eprinttype  = {arXiv},
  year      = {2018},
}

@Article{Ren2016,
  author        = {Ren, Shaoqing and He, Kaiming and Girshick, Ross and Sun, Jian},
  title         = {Faster R-CNN: towards real-time object detection with region proposal networks},
  journal       = {IEEE transactions on pattern analysis and machine intelligence},
  year          = {2016},
  volume        = {39},
  number        = {6},
  pages         = {1137--1149},
  __markedentry = {[jfnies:]},
  publisher     = {IEEE},
}

@InProceedings{Fang2017,
  author       = {Fang, Yuan and Kuan, Kingsley and Lin, Jie and Tan, Cheston and Chandrasekhar, Vijay},
  title        = {Object detection meets knowledge graphs},
  year         = {2017},
  organization = {International Joint Conferences on Artificial Intelligence},
}

@Article{Rueden2019,
  author      = {Laura von Rueden and Sebastian Mayer and Katharina Beckh and Bogdan Georgiev and Sven Giesselbach and Raoul Heese and Birgit Kirsch and Julius Pfrommer and Annika Pick and Rajkumar Ramamurthy and Michal Walczak and Jochen Garcke and Christian Bauckhage and Jannis Schuecker},
  title       = {Informed Machine Learning -- A Taxonomy and Survey of Integrating Knowledge into Learning Systems},
  date        = {2019-03-29},
  doi         = {10.1109/TKDE.2021.3079836},
  eprint      = {1903.12394v3},
  eprintclass = {stat.ML},
  eprinttype  = {arXiv},
  year      = {2019},
}

@InProceedings{Tong2006,
  author       = {Tong, Hanghang and Faloutsos, Christos and Pan, Jia-Yu},
  title        = {Fast random walk with restart and its applications},
  booktitle    = {Sixth international conference on data mining (ICDM'06)},
  year         = {2006},
  pages        = {613--622},
  organization = {IEEE},
}

@Article{Carion2020,
  author      = {Nicolas Carion and Francisco Massa and Gabriel Synnaeve and Nicolas Usunier and Alexander Kirillov and Sergey Zagoruyko},
  title       = {End-to-End Object Detection with Transformers},
  date        = {2020-05-26},
  eprint      = {2005.12872v3},
  eprintclass = {cs.CV},
  eprinttype  = {arXiv},
  year      = {2020},
}

@Article{lecun2015deep,
  title={Deep learning},
  author={LeCun, Yann and Bengio, Yoshua and Hinton, Geoffrey},
  journal={nature},
  volume={521},
  number={7553},
  pages={436--444},
  year={2015},
  publisher={Nature Publishing Group UK London}
}

@Article{Speer2012,
  author    = {Speer, Robert and Havasi, Catherine and CHAIDEZ, J and VENEZUELA, J and KUO, Y},
  title     = {Conceptnet 5},
  journal   = {Tiny Transactions of Computer Science},
  year      = {2012},
  publisher = {Citeseer},
}

@Article{Liu2021,
  author  = {Liu, Yan and Zhang, Zhijie and Niu, Li and Chen, Junjie and Zhang, Liqing},
  title   = {Mixed supervised object detection by transferring mask prior and semantic similarity},
  journal = {Advances in Neural Information Processing Systems},
  year    = {2021},
  volume  = {34},
}

@InProceedings{Menglong2019,
  author       = {Menglong, Cui and Detao, Ji and Ting, Zeng and Dehai, Zhang and Cheng, Xie and Zhibo, Chen and Xiaoqiang, Xia},
  title        = {Image classification based on image knowledge graph and semantics},
  booktitle    = {2019 IEEE 23rd International Conference on Computer Supported Cooperative Work in Design (CSCWD)},
  year         = {2019},
  pages        = {81--86},
  organization = {IEEE},
}

@Article{Zou2019,
  author      = {Zhengxia Zou and Zhenwei Shi and Yuhong Guo and Jieping Ye},
  title       = {Object Detection in 20 Years: A Survey},
  date        = {2019-05-13},
  eprint      = {1905.05055v2},
  eprintclass = {cs.CV},
  eprinttype  = {arXiv},
  year      = {2019},
}

@Article{Cordts2016,
  author     = {Marius Cordts and Mohamed Omran and Sebastian Ramos and Timo Rehfeld and Markus Enzweiler and Rodrigo Benenson and Uwe Franke and Stefan Roth and Bernt Schiele},
  title      = {The Cityscapes Dataset for Semantic Urban Scene Understanding},
  journal    = {CoRR},
  year       = {2016},
  volume     = {abs/1604.01685},
  bibsource  = {dblp computer science bibliography, https://dblp.org},
  eprint     = {1604.01685},
  eprinttype = {arXiv},
  url        = {http://arxiv.org/abs/1604.01685},
}

@inproceedings{optuna_2019,
    title={Optuna: A Next-generation Hyperparameter Optimization Framework},
    author={Akiba, Takuya and Sano, Shotaro and Yanase, Toshihiko and Ohta, Takeru and Koyama, Masanori},
    booktitle={Proceedings of the 25th {ACM} {SIGKDD} International Conference on Knowledge Discovery and Data Mining},
    year={2019}
}

@Misc{Lemmens2021,
  author    = {Lemmens, Jarl and Jancura, Pavol and Dubbelman, Gijs and Elforai, Hala},
  title     = {[Re] Object Detection Meets Knowledge Graphs - Supporting Datasets},
  year      = {2021},
  copyright = {Creative Commons Attribution 4.0 International},
  doi       = {10.5281/ZENODO.7385730},
  publisher = {Zenodo},
}

@article{castellano2022leveraging,
  title={Leveraging knowledge graphs and deep learning for automatic art analysis},
  author={Castellano, Giovanna and Digeno, Vincenzo and Sansaro, Giovanni and Vessio, Gennaro},
  journal={Knowledge-Based Systems},
  volume={248},
  pages={108859},
  year={2022},
  publisher={Elsevier}
}

@Misc{Beemelamanns2022,
  author = {Till Beemelmanns},
  title  = {cityscapes to coco conversion},
  month  = {07},
  year   = {2022},
  url    = {https://github.com/TillBeemelmanns/cityscapes-to-coco-conversion},
}

@article{zhu2022multi,
  title={Multi-modal knowledge graph construction and application: A survey},
  author={Zhu, Xiangru and Li, Zhixu and Wang, Xiaodan and Jiang, Xueyao and Sun, Penglei and Wang, Xuwu and Xiao, Yanghua and Yuan, Nicholas Jing},
  journal={IEEE Transactions on Knowledge and Data Engineering},
  year={2022},
  publisher={IEEE}
}

@article{Zhu2021SemanticRR,
  title={Semantic Relation Reasoning for Shot-Stable Few-Shot Object Detection},
  author={Chenchen Zhu and Fangyi Chen and Uzair Ahmed and Zhiqiang Shen and Marios Savvides},
  journal={2021 IEEE/CVF Conference on Computer Vision and Pattern Recognition (CVPR)},
  year={2021},
  pages={8778-8787}
}

@article{MacAodha2019PresenceOnlyGP,
  title={Presence-Only Geographical Priors for Fine-Grained Image Classification},
  author={Oisin Mac Aodha and Elijah Cole and Pietro Perona},
  journal={2019 IEEE/CVF International Conference on Computer Vision (ICCV)},
  year={2019},
  pages={9595-9605}
}

@article{redmon2018yolov3,
  title={Yolov3: An incremental improvement},
  author={Redmon, Joseph and Farhadi, Ali},
  journal={arXiv preprint arXiv:1804.02767},
  year={2018}
}
\appendix
\section{Appendix}
The results of the experiments using a value of $\epsilon > 1$ are collected here. Note that the baseline is unaffected by this and is, therefore, identical to the values given in the main part of this report. As some of these configurations were unable to achieve any mAP gain on FRCNN, values exhibiting a larger increase in recall were instead selected so as to show some effect of the model. However, as these effects are marginal in size and lack a robust mathematical foundation, we must caution against placing an excessive burden of interpretation on them. 

\subsection{FRCNN}

\begin{table}[H]
    \centering
    \begin{tabular}{|c|c c|c c c|}
        \hline
        Re-Op     &  mAP   & Recall & Small & Medium & Large \\
        \hline
        Baseline  &  25.35        & 35.60         & 12.26 & 34.77 & 57.43 \\
        KG        &  24.94 (-0.4) & 35.68 (+0.08) & 12.05 & 35.06 & 57.60 \\
        Hybrid    &  25.27 (-0.07)& 35.63 (+0.03) & 12.19 & 34.88 & 57.43 \\
        Freq.     &  25.38 (+0.02)& 35.59 (-0.01) & 12.26 & 34.75 & 57.44\\
        \hline
        
    \end{tabular}
    \bigskip
    \caption{Effects of different re-optimization methods on the FRCNN model's performance, with $\epsilon>0$. Numbers in bracket give the difference from the baseline.}

\end{table}

\begin{table}[H]
    \label{classwise1}
    \centering
    \begin{tabular}{|c| c c c c |c c c c|}
         \hline
         \multirow{2}{*}{Method} & \multicolumn{2}{c||}{AP} & \multicolumn{2}{c|}{Recall} \\
         \cline{2-3} \cline{4-5}
                        & Baseline & Freq & Baseline & Freq \\
         \hline 
         Person         & 25.29 & 25.31 & 33.51 & 33.54 \\
         Rider          & 29.76 & 29.83 & 41.31 & 41.25 \\
         Car            & 44.28 & 44.33 & 49.87 & 49.87 \\
         Truck          & 19.36 & 19.40 & 31.51 & 31.51 \\
         Bus            & 36.83 & 36.83 & 46.73 & 46.73 \\
         Train          & 12.14 & 12.13 & 23.48 & 23.48 \\
         Motorcycle     & 14.87 & 15.0  & 27.18 & 27.18 \\
         Bicycle        & 20.28 & 20.23 & 31.21 & 31.19 \\
         \hline
         
    \end{tabular}
    \bigskip
    \caption{Class-wise breakdown of Precision and Recall using Hybrid Re-Optimization with an FRCNN backbone, with $\epsilon>1$}

\end{table}

\subsection{DETR}

\begin{table}[H]
    \centering
    \begin{tabular}{|c|c c|c c c|}
        \hline
        Re-Op     &  mAP   & Recall & Small & Medium & Large \\
        \hline
        Baseline  & 21.84         & 35.48         & 17.46 & 44.63 & 62.71 \\ 
        Hybrid    & 22.08 (+0.24) & 35.50 (+0.02) & 17.42 & 44.51 & 63.03 \\
        KG        & 21.81 (+0.51) & 35.00 (-0.48) & 17.15 & 44.06 & 62.45 \\ 
        Freq.     & 22.19 (+0.35) & 35.36 (-0.11) & 17.36 & 44.37 & 62.82 \\ 
        \hline
        
    \end{tabular}
    \bigskip
    \caption{Effects of different re-optimization methods on the DETR model's performance. Numbers in bracket give the difference from the baseline.}
   
\end{table}

\begin{table}[H]
    \label{classwise1}
    \centering
    \begin{tabular}{|c| c c c c |c c c c|}
         \hline
         \multirow{2}{*}{Method} & \multicolumn{4}{c||}{AP} & \multicolumn{4}{c|}{Recall} \\
         \cline{2-4} \cline{5-9}                   
                      & Baseline & KG & Hybrid & Freq & Baseline & KG & Hybrid  & Freq \\
         \hline
         Person       &  28.58  & 28.54 & 28.65 & 28.40 & 40.75 & 40.10 & 40.69 & 40.53\\
         Rider        &  27.53  & 27.59 & 27.68 & 27.74 & 37.22 & 36.94 & 37.19 & 37.13\\
         Car          &  49.03  & 48.95 & 49.08 & 49.01 & 57.66 & 57.23 & 57.65 & 57.52\\
         Truck        &  11.88  & 11.88 & 12.19 & 12.32 & 29.73 & 28.51 & 29.77 & 29.27\\
         Bus          &  17.13  & 17.04 & 17.65 & 18.33 & 29.25 & 29.40 & 29.42 & 29.42\\
         Train        &   2.52  &  2.55 &  2.77 &  3.94 & 23.77 & 23.19 & 23.91 & 23.84\\
         Motorcycle   &  14.70  & 14.62 & 14.96 & 15.44 & 26.76 & 26.74 & 26.76 & 26.74\\
         Bicycle      &  23.36  & 23.33 & 23.69 & 23.71 & 38.71 & 37.88 & 38.65 & 38.49\\
         \hline
         
    \end{tabular}
    \bigskip
    \caption{Class-wise breakdown of Precision and Recall using Hybrid Re-Optimization with a DETR backbone}

\end{table}

\end{document}